\documentclass{article}
\usepackage[utf8]{inputenc}

\usepackage{lineno}
\usepackage{hyperref}
\usepackage{url}
\usepackage{tabularx}
\usepackage[svgnames,table]{xcolor}
\usepackage{amsmath}
\usepackage{amsfonts}
\usepackage{parskip}
\usepackage{tcolorbox}
\usepackage[numbers,sort&compress]{natbib}
\usepackage{rotating}
\usepackage{bm}
\usepackage{titling}

\usepackage{ragged2e}
\usepackage{url}

\newtheorem{definition}{Definition}

\title{Synthetic Data - what, why and how?}

\date{}

\newcounter{noteJJctr} 

\definecolor{colour4}{RGB}{51,160,44}
\newcounter{noteGCctr} 

\newcounter{noteMCctr} 

\newcounter{noteAEctr} 

\newcounter{noteLSctr} 

\definecolor{colourFH}{RGB}{153, 102, 51}
\newcounter{noteFHctr} 

\newcounter{noteCMctr} 

\definecolor{colourSC}{RGB}{51,180,144}
\newcounter{noteSCctr} 

\pretitle{\LARGE \begin{center}
}

\posttitle{
  \par\vspace{1cm}
  \large
    \begin{tabular}{c@{\extracolsep{4em}}c}
         James Jordon &  Lukasz Szpruch\\
         jjordon@turing.ac.uk & l.szpruch@ed.ac.uk \\
         & \\
         Florimond Houssiau & Mirko Bottarelli \\
         fhoussiau@turing.ac.uk & mirko.bottarelli@warwick.ac.uk \\
         & \\
         Giovanni Cherubin & Carsten Maple \\
         gcherubin@turing.ac.uk & cm@warwick.ac.uk \\
         & \\
         Samuel N. Cohen & Adrian Weller \\
         scohen@turing.ac.uk & aweller@turing.ac.uk \\
    \end{tabular}
  \par\vspace{1cm}
  \includegraphics[height=2.5cm]{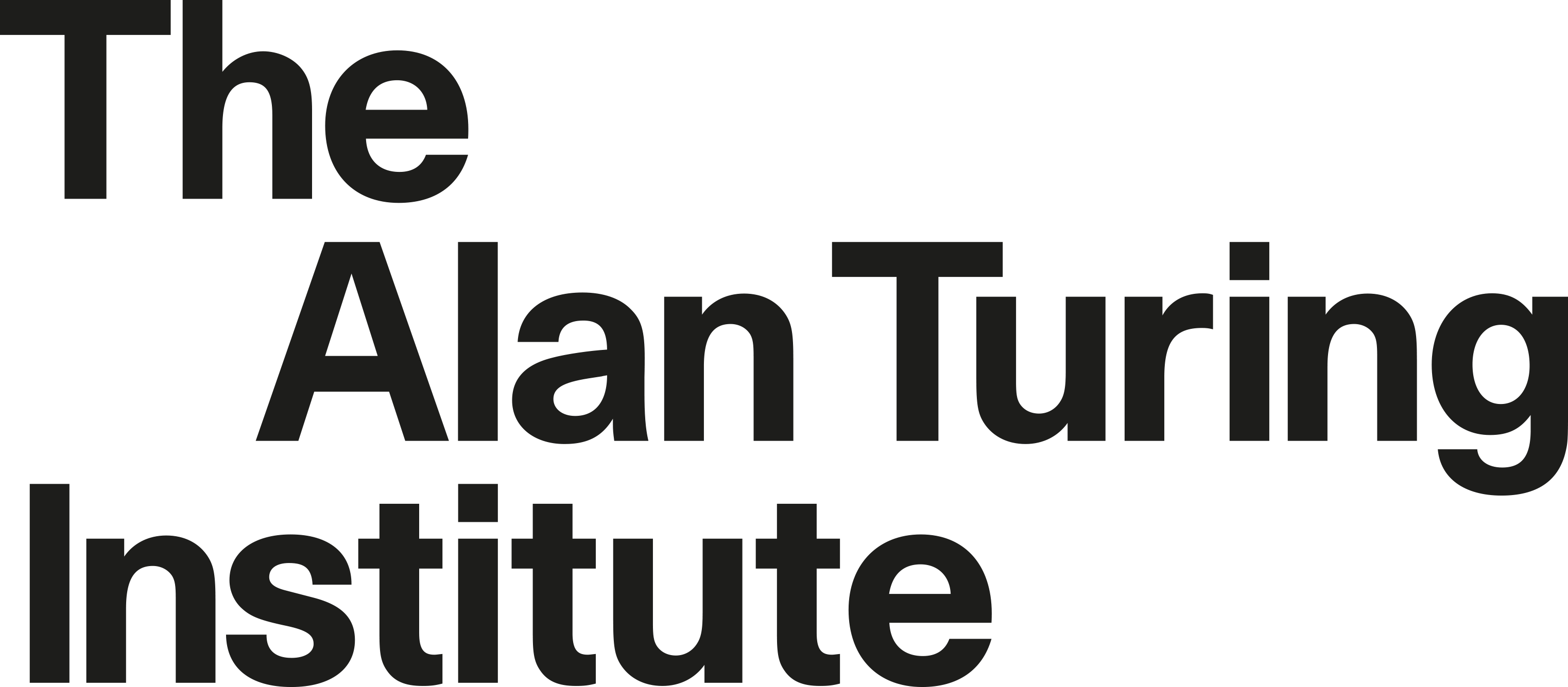}\quad\quad
  \includegraphics[height=4.5cm]{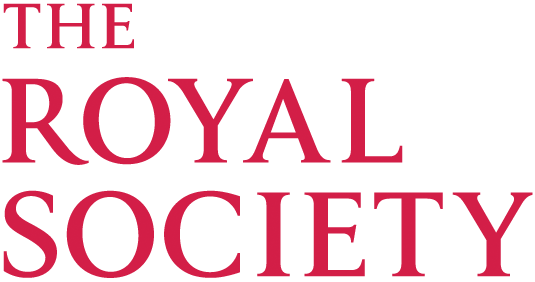}
  \end{center}
}
\preauthor{
}
\author{
\begin{tabular}{c@{\extracolsep{4em}}c}
     James Jordon &  Lukasz Szpruch\\
     jjordon@turing.ac.uk & l.szpruch@ed.ac.uk \\
     & \\
     Florimond Houssiau & Mirko Bottarelli \\
     fhoussiau@turing.ac.uk & mirko.bottarelli@warwick.ac.uk \\
     & \\
     Giovanni Cherubin & Carsten Maple \\
     gcherubin@turing.ac.uk & cm@warwick.ac.uk \\
     & \\
     Samuel N. Cohen & Adrian Weller \\
     scohen@turing.ac.uk & aweller@turing.ac.uk \\
\end{tabular}
}
\postauthor{
}

\begin{document}

\begin{titlepage}
    \begin{center}
        \vspace*{1cm}

        \LARGE
        \textbf{Synthetic Data - what, why and how?}

        \vspace{0.5cm}
        
        \vspace{1.5cm}

        \normalsize
        \large
        \begin{tabular}{c@{\extracolsep{4em}}c}
            James Jordon &  Lukasz Szpruch\\
            jjordon@turing.ac.uk & l.szpruch@ed.ac.uk \\
            & \\
            Florimond Houssiau & Mirko Bottarelli \\
            fhoussiau@turing.ac.uk & mirko.bottarelli@warwick.ac.uk \\
            & \\
            Giovanni Cherubin & Carsten Maple \\
            gcherubin@turing.ac.uk & cm@warwick.ac.uk \\
            & \\
            Samuel N. Cohen & Adrian Weller \\
            scohen@turing.ac.uk & aweller@turing.ac.uk \\
        \end{tabular}

        \vfill

        %\colorbox{black}{\includegraphics[height=2.5cm]{images/atilogo.png}}
        \includegraphics[height=2.5cm]{images/atiblack.png}
        \hspace{0.5cm}
        %\colorbox{black}{\includegraphics[height=2.5cm]{images/rslogo.png}}
        \includegraphics[height=2.5cm]{images/rslogo.png}
        \newline\par
        \vspace*{3cm}
        This report was commissioned by the Royal Society.
    \end{center}
\end{titlepage}
%\maketitle

\clearpage
\section*{Executive Summary}
This explainer document aims to provide an overview of the current state of the rapidly expanding work on synthetic data technologies, with a particular focus on privacy. The article is intended for a non-technical audience, though some formal definitions have been given to provide clarity to specialists. This article is intended to enable the reader to quickly become familiar with the notion of synthetic data, as well as understand some of the subtle intricacies that come with it. We do believe that synthetic data is a very useful tool, and our hope is that this report highlights that, while drawing attention to nuances that can easily be overlooked in its deployment.

The following are the key messages that we hope to convey. 

\paragraph{Synthetic data is a technology with significant promise.} There are many applications of synthetic data: privacy, fairness, and data augmentation, to name a few. Each of these applications has the potential for a tremendous impact but also comes with risks. 

\paragraph{Synthetic data can accelerate development.} Good quality synthetic data can significantly accelerate data science projects and reduce the cost of the software development lifecycle. When combined with secure research environments and federated learning techniques, it contributes to data democratisation.

\paragraph{Synthetic data is not automatically private.} A common misconception with synthetic data is that it is inherently private. This is not the case. Synthetic data has the capacity to leak information about the data it was derived from and is vulnerable to privacy attacks. Significant care is required to produce synthetic data that is useful and comes with privacy guarantees.  

\paragraph{Synthetic data is not a replacement for real data.} Synthetic data that comes with privacy guarantees is necessarily a distorted version of the real data. Therefore, any modelling or inference performed on synthetic data comes with additional risks. It is our belief that synthetic data should be used as a tool to accelerate the ``research pipeline'' but, ultimately, any final tools (that will be deployed in the real world) should be evaluated, and if necessary, fine-tuned, on the real data.

\paragraph{Outliers are hard to capture privately.} Outliers and low probability events, as are often found in real data, are particularly difficult to capture and include in a synthetic dataset in a private way. For example, it would be very difficult to ``hide'' a multi-billionaire in synthetic data that contained information about wealth. A synthetic data generator would either not accurately replicate statistics regarding the very wealthy or would reveal potentially private information about these individuals.

\paragraph{Empirically evaluating the privacy of a single dataset can be problematic.} Rigorous notions of privacy (e.g differential privacy) are a requirement on the {\em mechanism that generated} a synthetic dataset, rather than on the dataset itself. It is not possible to rigorously evaluate the privacy of a given synthetic dataset by directly comparing it with real data. Empirical evaluations can prove useful as tools to detect possible flaws in an algorithm or its implementation but may lead to false claims of privacy when there is none.

\paragraph{Black box models can be particularly opaque when it comes to generating synthetic data.} Overparametrised generative models excel in producing high-dimensional synthetic data, but the levels of accuracy and privacy of these datasets are hard to estimate and can vary significantly across produced data points.

\paragraph{Synthetic data goes beyond privacy.} Synthetic data provides promising tools to improve fairness, bias and the robustness of machine learning systems, but significantly more research is required to fully understand the opportunities and the limitations of this approach.

\clearpage

{
  \tableofcontents
}

\section{Introduction}

The availability of high volume, high velocity and high variety datasets, together with advanced statistical tools for extracting information, has the potential to improve decision-making and accelerate research and innovation. At the same time, many large-scale datasets are highly sensitive (e.g. in health or finance) and sharing them may violate fundamental rights guarded by modern privacy regulations (e.g. GDPR or CCPA). A large number of real-world examples demonstrate that high-dimensional, often sparse, datasets are inherently vulnerable to privacy attacks and that existing anonymisation techniques do not provide adequate protection. This limits our ability to share these large datasets, creating a bottleneck on the development and deployment of machine learning and data science methods.

Synthetic data is generated by a model, often with the purpose of using it in place of real data. By controlling the data generation process, the end-user can, in principle, adjust the amount of private information released by synthetic data and control its resemblance to real data. As well as addressing privacy concerns, one can to adjust for biases in historical datasets and to produce plausible hypothetical scenarios.

If used responsibly, synthetic data promises to enable learning across datasets when the privacy of the data needs to be preserved; or when data is incomplete, scarce or biased. It can help researchers and developers prototype data-driven models and be used to verify and validate machine learning pipelines, providing some assurance of performance. It can also fuel responsible innovation by creating digital sandbox environments used by startups and researchers in hackathon-style events.

Each of these uses presents great opportunities, but also challenges that require tailor-made solutions. Synthetic data generation is a developing area of research, and systematic frameworks that would enable the deployment of this technology safely and responsibly are still missing.

\subsection{Report Structure}
This explainer is organised as follows. In Section \ref{sec:what} we introduce a definition for synthetic data, give a brief history of its inception, and begin to answer one of the core questions surrounding synthetic data: {\em can it replace replace real data?} In Section \ref{sec:app}, we introduce key machine learning applications for synthetic data. 

Sections \ref{sec:privacy}-\ref{sec:privsdg} are dedicated to private synthetic data generation. In Section \ref{sec:privacy}, we introduce privacy from a more general perspective than synthetic data, introducing differential privacy and discussing some of its limitations. In Section \ref{sec:3att} we discuss three key attributes for evaluation of private synthetic data: utility, fidelity, and privacy. In Section \ref{sec:privaudit}, we discuss empirical evaluations of synthetic data for these key attributes. In Section \ref{sec:privsdg}, we discuss key differences between general privacy, and privacy as applied to the generation of synthetic data. We also survey existing methods in the space of private synthetic data generation. At the end of the section we discuss partially synthetic data, in which synthetic data is generated to create a hybrid real-synthetic dataset.

In Sections \ref{sec:debias} and \ref{sec:dataaug} we discuss synthetic data for fairness and data augmentation. In Section \ref{sec:generalgen}, we provide a more in-depth survey of existing generative models. Finally, in Section \ref{sec:indmess}, we summarise key themes from discussions with industry partners and start-ups in the field of synthetic data.

\section{What is Synthetic Data?} \label{sec:what}

Despite tremendous interest in synthetic data, \cite{jordon2020synthetic,assefa2020generating,mendelevitch2021fidelity,bellovin2019privacy,mckenna2021winning}, to the best of our knowledge there is no widely accepted definition. In order to encapsulate the full breadth of applications and approaches to synthetic data, we propose the following definition. 

\begin{definition}
Synthetic data is data that has been generated using a purpose-built mathematical model or algorithm, with the aim of solving a (set of) data science task(s).
\end{definition}

We contrast synthetic data with real data, which is generated not by a model but by real world systems (e.g financial transactions, satellite images, medical tests etc.). The {\em  model} -- the synthetic data generator --  can take many forms, from deep learning architectures such as the popular Generative Adversarial Networks (GANs) \cite{gan}, or Variational Auto-encoders (VAEs) \cite{vae}, through agent-based and econometric models \cite{bonabeau2002agent}, to a set of (stochastic) differential equations modeling a physical or economic system \cite{carmona2018probabilistic}.

Using computer-generated synthetic data to solve particular tasks is not a new idea, and can be dated back at least as far as the pioneering work of Stanislaw Ulam and John von Neumann in the 1940s on Monte Carlo simulation methods. Synthetically generated data has been widely used in research, as it provides a `ground truth', which is very useful in developing and evaluating machine learning pipelines.

The recent increase in data protection regulations has fueled the use of synthetic data to mitigate disclosure risk. The key hope, which goes back to work by Rubin and Little, \cite{rubin1987multiple,rubin1993discussion,little1993statistical}, is to be able to use synthetic data in place of real data, to avoid privacy concerns \cite{dpbook, privbayes, dpganxie, rcgan, pategan}.

The need to extract actionable information from large datasets led to the development of complex, data-driven (machine learning) models. For these models, the role of data in driving model selection is more prominent than it is for simpler, handcrafted models. This means that the quality of the model's output is directly dependent on the quality of the data used to train these models. This leads to a number of uses for synthetically generated data.
One such use is bias removal (e.g. historical biases in gender or race) \cite{feldman2015certifying, kamiran2009classifying, zhang2016causal, calmon2017optimized, fairgan, decaf}. Given biased training data, a natural approach is to train models using available data; these biases can then be seen in the output of the trained models. Rather than attempting to de-bias each trained model individually, one could generate a de-biased synthetic dataset and use it to train each model \cite{fairgan, decaf}, creating a unified approach for handling biases across an organisation. Another use would be to use synthetic data to enlarge datasets that are too small, e.g. to provide robustness against ``outlier'' examples \cite{wong2016understanding, gansemisup, badgansemi, ganmanisemi, timesemi, timeressemi}. Another key use case, to which we pay particular attention in this report, is the goal of using synthetic data to protect privacy. In each of these cases, the goal is to create synthetic data which resembles some aspects of the real data but not others. To maximise the utility of synthetic data, a fine balance must often be struck between competing objectives.

It is crucial to understand that synthetic data does {\em not} automatically address any of these problems. Training an off-the-shelf generative model based on real data, and then using this trained model to generate synthetic data, is {\em not inherently private}. Standard GANs do not generate private nor unbiased data. In fact, machine learning models have demonstrated the capability to (undesirably) memorise their training inputs \cite{arpit2017closer, goodfellow2016deep}. Applied to GANs, this can result in memorisation and regurgitation of the training data \cite{modecollapse}, undermining privacy in the synthetic data. At the other extreme, synthetic data can be generated without training data, e.g. using agent-based models that mimic the data generation process, such as agents transacting in a financial network. With no access to any real data, the synthetic data generator is private, but the data it generates is limited to the model's predetermined configuration, and will not enable statistical inferences to be reliably drawn about the real world.

\subsection{Can synthetic data replace real data?} \label{sec:likerealdata}
The use of synthetic data raises two key questions:
\begin{enumerate}
    \item Can we do the same things {\em with} synthetic data that we do {\em with} real data?
    \item Can we do the same things {\em to} synthetic data that we do {\em to} real data?
\end{enumerate}
The sorts of things we may wish to do {\em with} synthetic data are building models, performing data analysis, testing hypothesis, etc. Things one may wish to do {\em to} synthetic data, for example, might be linking separate datasets together, or extending a synthetic dataset when new records are added to the original dataset.

\subsubsection{How should we approach doing things {\em with} synthetic data?}
Ideally, one may hope that synthetic data can be simply plugged in wherever one might usually use real data (e.g. as training data for a model). Many papers on synthetic data evaluate it in this way. However, with private data, a more careful approach may lead to more accurate information being extracted from the synthetic data \cite{williams2010probabilistic}. In particular, conclusions from data analysis and hypothesis testing are necessarily weaker when using synthetic rather than real data, and the statistical significance of such analyses needs to be adjusted accordingly.

A particular concern for private data is bias. Ghalebikesabi et al.~\cite{ghalebikesabi2021bias} warn against the risks of learning from synthetic data, and propose a methodology for learning unbiasedly from such data. Wilde et al. \cite{wilde2021foundations} demonstrate superior performance when model parameters are updated using Bayesian inference, rather than approaches that fail to account for the fact the training data is synthetic.

\subsubsection{Data Linking} \label{sec:datalinking}
Something that can naturally be done with real data is linking. One dataset may contain an individual's lab test results, another may contain their genetic data, and another their hospital appointments. Each of these datasets can be linked to create a larger dataset containing information about inter-dataset correlations. If these datasets were synthesised independently, the 1-1 match between datasets will be broken; if, in the future, someone wished to pull together these synthetic datasets to investigate the correlations between, say, genetic data and lab test results, they would not be able to do so effectively.

One solution would be to encourage data holders to generate synthetic data with other (previously generated) synthetic datasets in mind. This may be appropriate in some situations (e.g. when the two datasets are being held by the same data holder), but in general this will not be the case. Moreover, the initial privacy loss suffered by an individual present in both datasets will be greater than if synthetic data was generated independently. This is particularly inefficient when linking the datasets might not be important, or the benefits of doing so are unclear.

In these situations, there is a need to be able to link two independently generated synthetic datasets (given access to real data) in a minimally privacy-leaking way. One workaround would be to simply generate a new joint dataset from the newly-joined underlying real datasets; but this does not leverage the existing synthetic data. A less naive approach would be to conditionally generate one of the two synthetic datasets based on the other (existing) synthetic dataset. This is a reduction in privacy cost over generating from scratch, but still fails to leverage the second already-generated synthetic dataset.

\subsection{Combining Synthetic Data with Other Technologies} \label{sec:othertech}

\textbf{Secure Research environment.} Synthetic data, in particular with differential privacy, has a natural application within secure research environments, in which decreasingly private data can be accessed in increasingly more ``secure'' environments. Consider releasing a dataset with strong privacy guarantees initially, evaluating a range of machine learning methods, then selecting the top $N$ candidates, and giving them access to less private data. This can be repeated, giving a `tournament' of methods. This raises the question of how to generate a series of datasets $\mathcal{D}_1, ..., \mathcal{D}_n$ which decrease in privacy individually and when taken together (so any subset $\mathcal{D}_1, ..., \mathcal{D}_j$, $j < n$ is as private as the terminal dataset, $\mathcal{D}_j$).

\textbf{Federated learning.} Federated learning is an emerging technology that enables training across decentralised datasets without pooling these datasets together. This contrasts with traditional centralised machine learning techniques, where the data is uploaded to one server, as well as to classical decentralised approaches which often assume that local data samples are identically distributed. 
In federated learning, distributed data holders allow an algorithm to be run on their private data, and only the (possibly noisy) outputs are released, without giving direct access to the data. The challenge with it is that, without accessing the data first, it might not be clear what algorithm one should run. Developing an algorithm on private synthetic data samples and evaluating its utility on real (distributed) data seems a very promising approach to this data bottleneck.

\section{Why use Synthetic Data?} \label{sec:app}
Synthetic data is being used as a solution to a variety of problems in many domains.
Three key areas that are of particular interest in a machine learning context are: (i) private data release (Section \ref{sec:privacy}); (ii) data de-biasing and fairness (Section \ref{sec:debias}); and (iii) data augmentation for robustness (Section \ref{sec:dataaug}). Although these are the areas that appear to have the most promise, this list is not exhaustive. Before going into details for each of them, we outline the key ideas for these areas and some of the specific use cases (and non-use cases) below.

\subsection{Private Data Release}
The wide adoption of data-driven machine learning solutions as the prevailing approach to innovate has created a need to share data. Without access to quality data, scientists and developers cannot make meaningful progress. However, GDPR, HIPAA, and a host of privacy regulations require data on individuals not to be shared carelessly (rightly so). The result is typically a long series of ``jumping through hoops'' in an attempt to access the necessary data. Synthetic data offers a potential solution. 

\paragraph{Development of ML tools.}
In this use case, a data controller may wish to assess an ML group's ability to solve a problem, or perhaps even assess several groups simultaneously to select the best partner with which to develop a final solution. In order to avoid privacy concerns, they plan to share synthetic data with their potential partners. For synthetic data to be useful in this setting, model development that is performed on the synthetic data should lead to the same conclusions as if it were carried out on the real data. More concretely, if a researcher comparing two models on the synthetic data were to conclude that model A outperforms model B for a given task, then the same conclusion should be reached when testing both algorithms on the real data. This suggests that, though the synthetic data would need to share many statistical properties with the real data, one can imagine that there are some properties that would not affect these comparisons. Once a final group/model has been determined, it can be taken to the real data for testing, tuning or even a complete re-training.

\paragraph{Software testing.}
There is a significant appetite for vast amounts of test production data for both system testing and User Acceptance Testing. Synthetic data can remove the requirement of going through lengthy and repeated approvals (e.g GDPR) and sanitation processes and hence save significant time and effort in the development lifecycle.  In this setting, it is important that synthetic data used for software testing is semantically correct, but it need not necessarily be statistically correct. Mathematically, this amounts to learning the {\em support} of the distribution and relevant structural properties (e.g time-series data), but not necessarily the distribution itself. Naturally, by not requiring statistical accuracy, there is much more room for increased privacy (or increased utility at the same privacy level). This is one of the design principles of OpenSafely~\cite{williamson2020factors}, where practitioners are able to test and develop algorithms on dummy data before running them once on the real data.

\paragraph{Deploying private machine learning tools.}
Machine learning models are not inherently private. It is well known that neural networks have the capability to memorise training inputs. Membership inference attacks are possible against such networks~\cite{shokri2017membership} and, as such, privacy-enforcing training algorithms for machine learning models have been developed \cite{dpsgd, abadi2016deep, pate1, pate2}. An apparent alternative (to enforcing privacy during the training of a model) would be to generate private data and then train a model using this data. {\em Perhaps} one advantage of such an approach would be that a single private synthetic dataset could create a unified approach to privacy (within a single organisation, say), but we believe that the cost in utility would outweigh the potential ``simplicity'' of the approach in most applications.

While training the model on private synthetic data might be appealing, it has  limitations. Private data generation isn't often able to capture all of the statistical structure that might be important to develop accurate models.
As such we believe it to be more prudent to focus efforts on the privacy of trained models, rather than on trying to generate private data with which to train.

\subsection{De-biasing}
\paragraph{Reducing/removing bias.} When generating synthetic data, one can aim at producing samples that do not suffer from historical biases but are otherwise still statistically accurate. Such data can be used then for training `black box' ML pipelines, while mitigating the risk of historical biases being amplified \cite{tiwald2021representative}. Importantly, such data can be reused to train multiple models. This should be contrasted with the approach of correcting each trained model separately. The latter approach has an additional disadvantage, as it could lead to inconsistencies in the way 'fairness and bias' are treated within an organisation. It must be recognised, however, that employing such methods to remove bias from the data introduces additional model risks that need to be quantified and monitored. 

\paragraph{What-if-scenario generation.}
Adjacent to bias removal in datasets is the question of {\em causal modelling} -- i.e. asking the question ``What if?''. Synthetic data may allow us to explore data generated according to the same causal structure but adjusted distributions, or with different causal interventions placed on the data generating process.
One must be very careful to properly model causal relationships though, as causal modelling is sensitive to assumptions and is {\em not} the same as conditional generation. Indeed, the trustworthy deployment of data-driven models requires that these perform well in situations that differ from the real data. Of course, again, we stress that model risk is being introduced as generative models are being used to produce these new scenarios.

\subsection{Data Augmentation}
\paragraph{Data labelling.} Deep neural networks are state of the art technology in computer vision applications. However, training deep neural networks requires vast amounts of (correctly) labelled data, which is often costly to produce. Synthetically generated labelled data offers a cost-efficient solution to this challenge, and has already been adopted by industry \cite{nikolenko2019synthetic}. In this application, one trains a neural network on synthetic data with the intention to deploy it on real data. In general, privacy is not of primary concern in these applications, as the data is not being used to replace the real data but to be used alongside it.

\section{Privacy in Machine Learning - An Overview} \label{sec:privacy}
Privacy is an incredibly large field, with practitioners coming from a wide variety of domains. Here, we present our view on privacy within the context of machine learning. We take a ground-up approach, motivating privacy through the notion of an adversary that has the potential to cause harm should too much information about an individual be revealed to them.

Privacy is a fundamental human right and a prerequisite for freedom of thought and expression.
For this reason, a key requirement of privacy is the \textit{consent} of individuals to have their data collected.
This consent usually relies on the expectation that the collection of their data, and the subsequent release of information derived from it, will not cause them \textit{harm}.
For some types of information, such as an individual's name, address, and phone number, the potential for harm is clear. For others, the potential for harm is more subtle: e.g. it might be that an insurance company would increase the price of an individual's insurance premiums based on knowledge that the specific individual is a smoker, without explicitly requesting this information. This potential for harm is caused by the ability of an {\em adversary} to gain information about an individual due to the release of data, or more generally from output that is derived from the data (again, this could be a synthetic dataset, or some other algorithm output). 

On the other hand, there is a highly social aspect to working with data. Shared datasets that can be used to benchmark models create a community that promotes rapid advancement of technology (see, for example, the rapid progress made in image classification that was caused by the availability of the MNIST and CIFAR datasets \cite{mnist, cifar}). In general, researchers want data, and it may not need to be too accurate for them to start being able to work with it.

Since the late 1990s, much research has been carried out within the realm of privacy. Early on, privacy was often tied together with the notion of {\em anonymity}, which came in a variety of flavours, from basic name/address/birthday removal (``pseudonymisation''), to $k$-anonymity \cite{kanom} (which itself had several iterations \cite{ldiv, tclose, dpres}). Although these notions apply to data, they really only have meaning on the raw data itself, and moreover have been shown to be inadequate even for that \cite{kcrit, imdb}. These approaches were built to prevent against {\em known} attacks. More recently, privacy is being studied with a view to prevent against abstract {\em threat models} rather than specific instantiations of an attack.

\subsection{The Threat Model View} \label{sec:threat}

Threat model privacy attacks can be summarised into 3 types: membership inference; attribute inference; and reconstruction attacks.

\vspace{1em}
\begin{tabularx}{\linewidth}{lXX}
    \textbf{Threat} & \textbf{Attacker's knowledge of Targeted Individual} & \textbf{Attacker's goal} \\
    Membership inference & Partial/Entire record &
        Determine if Targeted Individual was in the original data\\
    Attribute inference & Partial record &
        Recover missing attributes of Targeted Individual's data\\
    Reconstruction attack & N/A & Recover entire records from the original data\\
\end{tabularx}

These attacks differ both in their goals and in some of the assumptions placed on the adversary's prior knowledge {\em of the targeted individual}. It should be noted, however, that these attacks do not necessarily place any assumptions on the prior knowledge that an adversary might have about the individuals it is {\em not} targeting. These attacks are not typically performed against a single individual but multiple individuals at once, with a breach of any individual's privacy being considered a success by the adversary.

\paragraph{Membership Inference.} Membership inference \cite{shokri2017membership, hu2021membership} aims to determine whether an individual (whose full record might be known to the adversary) was part of the data that was given as input to an algorithm, given the output of the algorithm (and, potentially, knowledge of the workings of the algorithm). On its own, this is only of relevance when membership in the dataset implies some information about an individual. For example, it is not particularly useful (or much of a privacy violation) to ascertain that a particular individual's data was in the 2021 UK Census dataset. Following the example of the smoker, though, it is a violation to be able to ascertain that a given individual is in a dataset that contains only smokers (such as a dataset used in a scientific study of smokers).

\paragraph{Attribute Inference.} Attribute inference \cite{al2012homophily, privtraits} is slowly becoming infamous as a non-violation of privacy.
The goal with attribute inference is to determine some extra information about an individual given some prior knowledge about some of their attributes and access to an algorithm's output (e.g. synthetic data, or a trained ML model). Of course, this is precisely what predictive models aim to do -- predict (an) attribute(s) from a set of other attributes \cite{kosinski2013private}. But this leads to an almost paradoxical conclusion, how can a model that was trained {\em without} an individual's data, violate their privacy? This almost-paradox has led to many privacy researchers abandoning attribute inference as a violation of privacy \cite{dpinf, scienceharm}.
However, the question should not be, ``does this allow you to learn more about an individual?'', but rather, ``does this allow you to learn more about an individual than if they had not been in the data?'' It is still possible (in fact, with ML it is very likely \cite{dlgen}) that a trained model will perform better on its training inputs than on inputs not used for its training. This indicates that the release of such a model does violate the privacy of the individuals in the training set \cite{aioverfit}. Recent work in the space has attempted to determine the feasibility of attribute inference attacks. \cite{zhao2021feasibility} show that even when membership inference is possible, attribute inference may not be; they do, however, demonstrate that {\em approximate} attribute inference is possible.

\paragraph{Reconstruction Attacks.} Reconstruction attacks~\cite{dinur2003revealing} aim to extract entire records from the training dataset, based on the output of an algorithm.
For instance, Dinur and Nissim showed how a database protected by a question-and-answer system can be reconstructed by an attacker if the level of the noise added to answers is low~\cite{dinur2003revealing}.
Another high profile example is the attack performed by researchers at the US Census bureau on aggregates from the 2010 Census. They were able to retrieve exact records for 46\% of the US population using publicly released data~\cite{uscensus2021}.
Unlike membership and attribute inference attacks, reconstruction attacks are not \textit{targeted}: they aim to retrieve records for any (or all) records in the original data. This could leverage prior knowledge about {\em some} of the individuals in the training set (with the remaining unknown individuals being considered the targets). 
Note that the same caveat as for attribute inference attacks applies: even an algorithm sampling records uniformly at random in $\mathcal{X}$ will reconstruct \textit{some} records in the real data with nonzero probability. Evaluation of a reconstruction attack should therefore be contrastive, i.e.~based on the difference of reconstruction likelihood for a record due to their presence in the data (although real-world attacks recover such large fractions of the data that this consideration is often not needed).

\subsection{Differential Privacy} \label{sec:dp}
Differential Privacy \cite{dpbook, nontechdp}, first proposed by Dwork et al.~in 2006 \cite{ogdp}, is becoming increasingly accepted as a robust, meaningful, and practical definition of privacy \cite{uscensus, microsoftdp, facebookdp}. 
Informally, differential privacy requires that an algorithm's (necessarily random) output not differ ``too much'' between {\em adjacent} datasets. Intuitively, because the outcome cannot differ significantly, there cannot be too much ``information leakage'' from the dataset to the algorithm output. The definition rests on some notion of datasets being {\em adjacent} to each other, and can be used to capture the notion of an individual's data. This adjacency can be defined in different ways depending on the type of data structure. With so-called tabular data, adjacency between two datasets is typically defined to mean that one can be obtained from the other by either the removal/addition (unbounded differential privacy \cite{dpbook}) or replacement (bounded differential privacy \cite{dpbook}) of a row.

Defining this adjacency amounts to deciding precisely what information should be protected \cite{mcsherry2009differentially, hardt2012beating}. With graph-like data, one could consider either entire nodes (along with all associated edges) to be important, or instead consider only the edges themselves to each be individually important \cite{edgenodedp, edgenodedp2}. With tabular data, one might define adjacency as datasets differing in precisely one value, allowing ``feature-wise'' differential privacy that protects each value contributed to the dataset, rather than rows as a whole. This would allow data contributors to decide to maintain the privacy of some, but potentially not all, of their data.

A particularly important feature of differential privacy is that it is {\em contrastive} -- it compares the outcome of an algorithm when an individual is in the training data to the outcome when the individual is not in the training data, or some similar adjacent perturbation. This idea, that privacy cannot be breached when an individual is not in the data, is crucial in dismissing several more ad-hoc notions of privacy. Crucially, for synthetic data, just because one of the synthetic data points {\em looks like} one of the original data points does not mean that privacy has been violated -- the synthetic point might have been generated even without the original point being present in the training data.

\begin{definition}[Differential Privacy \cite{dpbook}]
\sloppy
	A {\em randomized} algorithm, $\mathcal{M}$, is $(\varepsilon,\delta)$-differentially private if for all $\mathcal{S} \subset \mathrm{Im}(\mathcal{M})$ and for all neighboring datasets $\mathcal{D}, \mathcal{D}'$:
	\begin{equation*}
	\mathbb{P}(\mathcal{M(\mathcal{D})} \in \mathcal{S}) \leq e^\varepsilon \mathbb{P}(\mathcal{M(\mathcal{D}')} \in \mathcal{S}) + \delta
	\end{equation*}
When $\delta = 0$, $\mathcal{M}$ is said to be \emph{pure $\varepsilon$-differentially private}.
\end{definition}

\fussy
Intuitively, the key promise of differential privacy is that \textit{any} analysis run on the output of a differentially private procedure will yield approximately the same result whether or not any individual contributes their record to the dataset.
This also includes potential harms that could be caused by the publication of potentially sensitive information.
For instance, assume that a DP procedure is used to train a ML model to detect a specific disease from sensitive medical records.
No operation performed on this model (e.g. inspecting its parameters, applying it to well chosen inputs) can reveal information about individual training records.
Hence, it serves as a form of statistical guarantee for individuals that the collection and use of their data will not yield negative consequences (that would not otherwise occur even if the data was not shared\footnote{A notorious~\cite{dpbook} example of this is an insurance company updating premiums based on the result of a study showing correlation between smoking and lung cancer. A client's premium might go up even if their data is not used in the study. Differential privacy here ensures that the result of the study would not change too much whether they give their data or not.}).
Formally, from a Bayesian point of view, this means that for all potential priors over datasets, the posterior computed after observing the outcome will be similar to the posterior obtained if any one user was removed from the dataset~\cite{kasiviswanathan2014semantics}.

\paragraph{Bayesian interpretation of Differential Privacy}
{\em It is instructive to consider the Bayesian interpretation of privacy guarantees implied by differential privacy, which compares the adversary’s prior with the posterior. To that end it is useful to view $\mathcal D$ and $\mathcal D'$ as a realisation of a random variable $\bm{ \mathcal D}$. That way we can model prior knowledge an adversary has about the dataset i.e $\mathbb P(\bm{ \mathcal  D} = \mathcal D)$. Note that $(\epsilon,0)$-differential privacy implies a bound on the Bayes factor
\[
\frac{\mathbb P( \mathcal M(\bm{ \mathcal D}) | \bm{ \mathcal D} = \mathcal D )}{\mathbb P( \mathcal M(\bm{ \mathcal D}) | \bm{ \mathcal D} = \mathcal D' )} \leq e^{\epsilon}\,.
\]
This then implies a privacy guarantee on posterior beliefs regarding the value of $\bm{\mathcal D}$ given the output of a differentially private algorithm
\[
\frac{\mathbb P( \bm{ \mathcal D} = \mathcal D | \mathcal M(\bm{ \mathcal D}))}
{\mathbb P( \bm{ \mathcal D} = \mathcal D' | \mathcal M(\bm{ \mathcal D}))}
=
\frac{\mathbb P( \mathcal M(\bm{ \mathcal D}) | \bm{ \mathcal D} = \mathcal D )}{\mathbb P( \mathcal M(\bm{ \mathcal D}) | \bm{ \mathcal D} = \mathcal D' )} 
\frac{\mathbb P( \bm{ \mathcal D} = \mathcal D ) }
{\mathbb P( \bm{ \mathcal D} = \mathcal D' ) } 
\leq e^{\epsilon}
\frac{\mathbb P( \bm{ \mathcal D} = \mathcal D )}
{\mathbb P( \bm{ \mathcal D} = \mathcal D') }\,. 
\]
To put it another way, and due to symmetry between $D$ and $D'$, differential privacy implies that the log-odds cannot change significantly,
\[
\left| \log \left(\frac{\mathbb P( \bm{ \mathcal D} = \mathcal D | \mathcal M(\bm{ \mathcal D}))}
{\mathbb P( \bm{ \mathcal D} = \mathcal D' | \mathcal M(\bm{ \mathcal D}))} \right)
- \log \left( \frac{\mathbb P( \bm{ \mathcal D} = \mathcal D )}
{\mathbb P( \bm{ \mathcal D} = \mathcal D') } \right)\right| \leq \epsilon.
\]}

In light of the threat model view introduced in Section \ref{sec:threat}, it should be noted that differential privacy provides provable bounds on the ability of an adversary to perform such attacks \cite{dpboundmi}. These bounds assume worst-case prior information, knowledge of the algorithms, and computing power of the adversary. Differential Privacy also enjoys several nice properties that have helped with its adoption, such as composability, resistance to post-processing, and plausible deniability \cite{dpbook}. Composability and the post-processing theorem, in particular, allow for differentially private algorithms to be built out of smaller building blocks, which is precisely the driving force behind Differentially Private Stochastic Gradient Descent (DPSGD) \cite{dpsgd, abadi2016deep}. DPSGD has enabled differential privacy to be applied to deep learning architectures. This, in turn, has allowed the development of several DPSGD-driven generative models for synthetic data (see Sec. \ref{sec:privmodels}).

\paragraph{Flavours of Differential Privacy.} Above, we introduced the two most common notions of differential privacy, pure- and approximate-differential privacy ($\varepsilon$ and $(\varepsilon, \delta)$). There are in fact several relaxations of differential privacy, such as Renyi-Differential Privacy \cite{renyidp}; extensions, such as Label-DP \cite{labeldp}; and even stronger notions, such as local differential privacy \cite{localdp}. Depending on the task at hand, these notions may be more or less useful than vanilla DP.

\subsubsection{Limits of DP} \label{sec:limitsdp}
Despite being widely accepted as the best available privacy definition, differential privacy is not without its weaknesses. 

\paragraph{Choosing parameters $\varepsilon, \delta$.}
The privacy protection afforded by a differentially private mechanism is controlled by parameters $\varepsilon$ and $\delta$.
Choosing appropriate values for these parameters is notoriously difficult (in part due to their opaqueness in interpretability), and strongly depends on the context~\cite{lee2011much, hsu2014differential, abowd2019economic}.
This is further complicated by the fact that many methodologies lack a tight analysis of their privacy, leading to larger-than-necessary noise being injected into the system, hindering utility~\cite{stadler2020synthetic}.

\paragraph{Relaxations.}
Though often celebrated as a strength, the lack of assumptions placed on an adversary's knowledge and capabilities can lead to overly conservative computations, which hinder the utility of the output.
Researchers have proposed many relaxations of the original definition (which required $\delta = 0$)~\cite{desfontaines2019sok}.
However, the privacy guarantees of these can be hard to understand and model. Similarly to the parameters issue, researchers have started using attacks to compare mechanisms using different definitions of privacy~\cite{labeldp}.

\section{Utility, Fidelity and Privacy of Synthetic Data.} \label{sec:3att}
For synthetic data to be {\em meaningful}, it must be similar to {\em and} different from the original data in some sense. If synthetic data is being considered, then there is a reason that the original data is inappropriate or inadequate for the task at hand --  be it because it is non-private, biased, or too small -- and so synthetic data that is too similar to the original data will also suffer from the same problems. The ``allowed'' similarity (or rather the required non-similarity) will differ from task to task, and constitutes one of the 3 attributes that are fundamental to synthetic data generation: {\em utility}, {\em fidelity}, and {\em privacy}.

\paragraph{Utility:} The utility of synthetic data often is determined by its usefulness for a given task or set of tasks.  This often involves contrasting the performance of models trained on real vs synthetic data, and might involve inspecting concrete metrics such as accuracy, precision, root mean-squared error, etc.; and/or model fairness properties such as demographic parity, fairness through unawareness, or conditional fairness \cite{decaf}. Doing so often requires the \emph{Train on Synthetic, Test on Real} (TSTR) paradigm \cite{rcgan} in which models are trained on synthetic data and their performance then evaluated on real data.

\paragraph{Fidelity:} Often lumped together with utility, we define fidelity to be measures that directly compare the synthetic dataset with the real one (rather than indirectly through a model, or through performance on a given task). From a high-level perspective, fidelity is how well the synthetic data ``statistically'' matches the real data. Measures of fidelity are often used because of an underlying intuition that a specific fidelity will correspond to improved performance on a wide range of tasks. In the most general case, full statistical similarity (i.e. matching the distributions of the synthetic and real data), should allow many tasks that would be performed on the real data to be performed on the synthetic. However, such a match is difficult, especially in the presence of privacy requirements \cite{ullman2011pcps}, and even undesirable in the presence of biases \cite{decaf}. Rather than seeking a ``full'' statistical match, one might inspect low-dimensional marginals \cite{choi2017generating}, the syntactical accuracy of the synthetic data \cite{alaa2021faithful}, or look at the distribution of the remaining features conditional on a feature that is known to be biased in the original data.

\paragraph{Utility vs. Fidelity:} 
Much of the literature on synthetic data, in particular for private synthetic data, focuses on the 2-dimensional trade-off between utility and privacy, folding fidelity into utility. While the two are unavoidably linked, they are not synonymous nor perfectly correlated. In some scenarios, fidelity can be reduced while leaving utility unaltered (or vice versa), potentially ``leaving room'' for other benefits, for example, improved privacy.

\paragraph{Privacy:}  The privacy of synthetic data is determined by the amount of information that it reveals about the real data used to produce it. Depending on the use case, different privacy guarantees might be required. For example, internal synthetic data release within a secure environment will typically require less stringent privacy evaluation than data released to the general public. Theoretically sound notions such as differential privacy and its offspring exist, allowing for systematic analysis of the privacy of algorithms used to produce synthetic data. Less is known about the precise meaning of the privacy of a specific synthetic data sample if the data generation method is not revealed, or how to evaluate it, since privacy is typically defined as a statistical property over many instances. Of course, extra care is required to ensure that privacy that has been proven on paper is not lost through sloppy implementation of these algorithms in practice \cite{stadler2020synthetic}.

\paragraph{Privacy vs Fidelity:} As a rule of thumb, when fidelity increases, the privacy of synthetic data decreases. This means that, in general, it is impossible to generate private synthetic data that will be useful for all use cases. Instead, one might group potential use cases in terms of the type of fidelity that is required (i.e.~which features of the original dataset need to be captured by the synthetic data) and generate multiple synthetic datasets, each with user specified privacy guarantees.

\subsection{Synthetic Data Desiderata}
A good synthetic data generator (SDG) should simultaneously satisfy the following properties:
\begin{enumerate}
    \item \textbf{Syntactical accuracy}: The generated data should be plausible (e.g.~a synthetically generated postcode should exist). However, this also requires that certain structural properties of the data are preserved. For example, with time-series data, one needs to ensure that data points are not generated using information from the future. Similarly, when synthesising financial transnational networks, the underlying graph structure of the data must be preserved. 
    \item \textbf{Privacy}: It should be possible to precisely quantify how much information about the original data is revealed through the releasing of the synthetic sample. How exactly one measures privacy will depend on the specific task at hand. While differential privacy is one popular way of assessing the amount of information release through synthetic data generators, a different notion might be required when the data is sparse or one wants to move away from worst-case bounds. 
    \item \textbf{Statistical accuracy}: It should be possible to precisely quantify the statistical similarity (or lack thereof) between the synthetic and the original data. When measuring statistical accuracy, one might be interested in capturing certain marginal distributions and certain relationships between variables, but not others. A good synthetic data generator should allow for control over this.   
    \item \textbf{Efficiency}: The algorithm should scale well with the dimension of the data space (i.e. feature space). It is well known that, in general, approximation of distributions can suffer from the curse of dimensionality, and consequently sampling from unstructured distributions is an NP-hard problem. 
\end{enumerate}

While it its relatively straightforward to design algorithms for which a subset of these properties hold, there is currently no systematic framework for developing SDGs for which all 4 properties are satisfied simultaneously. For example, generation of statistically accurate but private data is hard, as these goals may be in conflict. Specifically within the realm of differentially private {\em synthetic data}, Ullman et al.~\cite{ullman2011pcps} demonstrated that a computationally efficient algorithm (i.e. runs in polynomial time) that generates synthetic data that both: (i) satisfies differential privacy; and (ii) preserves the correlations between pairs of features, does not exist. This result holds in a general sense, in that for every algorithm that {\em could} generate synthetic data, there is {\em a} dataset that ``will not work''. That said, it is possible that:
\begin{itemize}
    \item for a specific application (e.g., dataset), it is possible to efficiently generate DP synthetic data;
    \item one may not be interested in the correlations being preserved (i.e. the application may demand a different fidelity notion).
\end{itemize}
Nevertheless, this impossibility result implies that one needs to assess the privacy and fidelity of the data on a per-case basis. Most importantly, there is no ``one-size-fits-all'' differentially private synthetic data generation method.

In \cite{boedihardjo2021covariance}, the authors show that by reducing the requirement that all correlations be matched to the requirement that {\em most} correlations be matched, a computationally efficient algorithm {\em does} exist. Such a result is promising for synthetic data, but raises the question of being able to quantify what aspects of the data structure (i.e. which correlations) are {\em not} being matched.

Such results highlight the need for synthetic data to not try to be too general -- synthetic data should be generated with a use case in mind. For a given use case, relevant statistical properties can be preserved, while others can be ignored in the name of creating privacy. Below, we give some concrete examples of such use cases, alongside an application we believe to be a misguided endeavour.

\section{Auditing Synthetic Data} \label{sec:privaudit}
In this section, we discuss various approaches for empirically evaluating synthetic data, both in terms of its privacy, and its utility and fidelity.

\subsection{Empirically Evaluating the Privacy of Synthetic Data} 
Given that differential privacy is a theoretical notion of privacy that must be proven, and correctly implemented,  to be satisfied, a natural question is to ask whether or not one can verify some notion of privacy for a synthetically generated dataset, or a synthetic data generator, empirically. Given that the goal is to protect against the (abstract) threat models outlined in \ref{sec:threat}, can one ``prove'' privacy by attempting to perform attacks against a given dataset?

\paragraph{DP verification.}
As mentioned above, the differential privacy of a synthetic dataset is more precisely a property of the algorithm that generated it, and {\em cannot} be verified by inspecting the synthetic dataset itself. Researchers have been investigating methods for checking that an algorithm meets DP requirements~\cite{ding2018detecting,jayaraman2019evaluating,jagielski2020auditing}. These methods work either by querying the algorithm in search of violations of the privacy definition, or by running known attacks (e.g.~membership inference) against it.

These are useful tools, which can be applied to SDG methods, as a way of testing/understanding their privacy. Perhaps the most useful application is to help us understand what values of $\varepsilon$ make the most sense in a specific context. However, using these tools to ``prove'' differential privacy is not possible, as they are based on statistical analysis of the generating algorithm. What is possible is that one can show, with a certain confidence, that an algorithm is {\em  likely} to be differentially private, but doing so would require sampling many, many times (and more samples would be needed for more complex outputs/algorithms, e.g.~when the output is a dataset) from the algorithm with many, many different input datasets. Doing so would be {\em highly} computationally intractable if any sort of meaningful level of confidence was to be achieved.

\paragraph{Leakage estimation.}
An alternative option for evaluating the privacy of algorithms is to use \textit{leakage estimation} techniques~\cite{chatzikokolakis2010statistical,cherubin2017bayes}, which stem from the quantitative information flow community~\cite{smith2009foundations}. These techniques enable quantifying the privacy of an algorithm with respect to a specific threat model (or adversary). For example, in the context of SDG, this means one could use leakage estimation for assessing the resilience of a method against membership inference or attribute inference attacks, which we described above.

A strong advantage of this approach, is that it does not require any formal analysis of the SDG method; additionally, it can be used for selecting the privacy parameters of a DP algorithm. Another advantage is that some of these methods enable a fully black-box analysis; that is, there is no need to describe the algorithm's internals analytically. One disadvantage is that the leakage estimation analysis is done with a specific threat model (or attack) in mind, although there are ways of capturing many attacks with the same analysis~\cite{m2012measuring}. A second disadvantage is that, in the case of black-box leakage estimation methods, the formal guarantees derived via these approaches make the assumption that we can sample an arbitrary amount of data from the algorithm. Nevertheless, they have been shown to be effective when tackling real-world tasks~\cite{cherubin2019fbleau}.

\paragraph{Empirical privacy evaluation of datasets themselves.} The empirical evaluation of privacy of synthetic data is a nascent and challenging area of research. Despite the fact that {\em differential} privacy cannot be established for a dataset in isolation, practitioners in the field of synthetic data have made use of a hold-out test set to evaluate (other notions of) the privacy of generated synthetic data \cite{mostlyai}. The Nearest-Neighbour distance ratio (NNDR) has been used to inspect whether or not synthetic data points are closer than some hold-out test points to the underlying real data points (on average). This involves splitting the data into a training set, and a test set (as is similarly done in supervised learning problems). The training set is then used to train the model. Once trained, samples are drawn from the trained model, and the distance from these samples to the training data is compared with the distance from the test set to the training data.

Note that although this method is agnostic of the method used to train the data generator, it does rely on a hold-out test set being available. As noted in \cite{mostlyaiblog}, non-existence of points can be just as revealing as the existence of points in the synthetic dataset, but privacy analysis via NNDR does not capture this behaviour. Indeed it is possible to satisfy NNDR by creating ``holes'' in the synthetic data around the real data points, but such an approach would reveal where the real records should be.

A similar application of NNDR is performed in \cite{adsgan} to attempt to ensure privacy of the generative model, which also does not control for the creation of such ``holes''. In \cite{choi2017generating}, they use a nearest-neighbour based classifier to quantify the risk of attribute disclosure but do so in a non-contrastive way, thus rendering the privacy analysis weak. An NNDR-type metric was also used in \cite{yale2019assessing} to assess the relative privacy of several generative models proposed for health data.

\paragraph{Attacks against private synthetic data.} As we explain in section~\ref{sec:threat}, one of the approaches to understand and analyse privacy is through the lens of attacks: what can a motivated attacker learn about users in the dataset?
In addition to being intuitive, this approach can help us evaluate whether a system protects user privacy in a given context, and compare methods built with different privacy definitions in mind.
This is particularly relevant for synthetic data generation: as we have presented in previous sections, a wide variety of SDG methods have been proposed, with widely different privacy definitions, choice of parameters, and assumptions.
Despite the potential of adversarial approaches to evaluate privacy risks of synthetic data, the development of privacy attacks against SDG remains underexplored. We here review existing attacks, and suggest promising research areas.

The main method to evaluate privacy risks in synthetic data was proposed by Stadler et al.~\cite{stadler2020synthetic} in a recent paper.
They propose a general methodology to apply membership and attribute inference attacks on \textit{any} synthetic data generation model.
They assume black-box access to the SDG method, and specifically, being able to retrain the SDG model on new data.
Indeed, analysing the synthetic data alone (as in NNDR metrics) is, in general, not sufficient to properly understand information leakages: an algorithm sampling records uniformly at random might, by coincidence, replicate exactly some records from some private dataset, but this would not usually be considered a privacy violation.
Further, this assumption is a key transparency requirement: in order to audit synthetic data, it is necessary to be able to understand how it was generated. Therefore, privacy guarantees cannot usually be obtained by maintaining secrecy of the generating algorithm.
The method proposed by Stadler et al. uses \textit{shadow modelling}:  the attacker simulates many runs of the SDG algorithm, using auxiliary data, to generate synthetic datasets trained with or without a target user. A binary classifier is then trained on features extracted from these synthetic datasets to predict whether the target user is in the training dataset.
Empirical results suggest that current SDG methods either are vulnerable to this attack or, if the attack fails, lead to an accuracy worse than non-SDG methods for a range of data analysis tasks.

Outside of this specific paper, there is a rich literature on privacy attacks that can be leveraged to develop attacks against synthetic data.
We here detail two possible lines of research for this approach:
\begin{enumerate}
    \item Many synthetic data generation methods rely on Generative Adversarial Networks (see, e.g., \cite{pategan, dpganxie, gpate, timegan, decaf}).
    Researchers have demonstrated that such GANs can be vulnerable to white-box and black-box membership inference attacks~\cite{chen2020ganleaks, logan}.
    A key question is then: how can these attacks be ported to the setup where the attacker has access to \textit{data} generated by the model, rather than the model itself.
    \item Some methods built with specific use cases in mind can aim to closely replicate statistical properties of the training dataset, such as one-way marginals histograms or correlations.
    Many membership inference attacks have been proposed  against aggregate statistics, from simple statistical tests~\cite{homer2008resolving,dwork2015robust} to advanced attacks based on shadow models~\cite{pyrgelis2017knock}.
    If the synthetic data accurately reproduces many aggregates from the original data, one can apply these attacks to the synthetic data to infer membership of specific records in the training dataset.
    This leads to an interesting question: how many statistics can be accurately reproduced from the original data, without enabling such attacks?
\end{enumerate}
These are only two prospective research directions for the adversarial evaluation of synthetic data, which is an open line of research.

\subsection{Evaluating the Utility and Fidelity of Synthetic Datasets} \label{sec:utilfidel}
The methods presented in this section generally focus on privacy as their primary design goal, most often through explicit guarantees such as differential privacy.
In this section, we review approaches to evaluate the secondary goal of such datasets: their \textit{utility} and \textit{fidelity}.
The utility of a private synthetic dataset is determined entirely by its application.
Generating synthetic data to enable release of otherwise private data has almost as many use cases as there are machine learning problems -- any data-driven problem might be derived from sensitive data and the data controllers may wish to investigate which ML methods might address the problem. 
Below (\ref{sec:privutil}) we give example use-cases, and suggest how utility might be measured in such cases.
Fidelity is less well-defined: it generally aims at evaluating how close the \textit{distribution} of the synthetic dataset is to that of the real data, the idea being that if the distributions match, the synthetic data can be used to perform any task as accurately as with the real data.
We discuss this more general use case, and review works studying fidelity, in section~\ref{sec:privfid}.

\subsubsection{Utility-driven evaluation} \label{sec:privutil}
The key use-case for privately-generated synthetic data is to enable research and industry data analysis tasks without access to sensitive data.
A particularly important application is the development of machine learning (ML) models to perform inference and classification tasks from data.
In this setting, the goal is to determine the best model (or a selection of best contenders), and train it (choose its parameters) to perform a given task.
In general, such a task will come with its own metric of performance (e.g. accuracy/AUROC in a classification task).
There are broadly two directions of research aiming to evaluate the suitability of synthetic data for ML training: (1) evaluating the performance of models trained on synthetic data, and (2) evaluating whether the \textit{relative performances} of different models are similar on synthetic and real data.

The first approach assumes that analysts will train a machine learning model on the synthetic data, and use this model directly on real, future data.
In this situation, it is important that the accuracy of a model estimated with synthetic data reflects its accuracy on real data.
For instance, Beaulieu et al. evaluate their synthetic data generation method by measuring the accuracy of classifiers trained on synthetic datasets on the real sensitive medical data used to generate the synthetic data~\cite{beaulieu2019privacy}.
Patki et al. pushed this further, by distributing synthetic datasets and real datasets randomly to teams of data scientists, and evaluating whether teams working on real and synthetic datasets would arrive at approximately the same conclusions~\cite{patki_synthetic_2016}.
Similar approaches were used by Tao et al.~\cite{tao2021benchmarking}, where a XGBoost classifier is trained on synthetic data and evaluated on real data for a range of different tabular datasets.

Note that this approach makes some assumptions on the family of models that will be trained, since it is impossible to test \textit{all possible classes of models}, as well as all possible choices of hyper-parameters.
When synthetic data is generated with a set of specific tasks in mind, custom metrics can also be developed that capture the accuracy on these specific tasks.
For instance, in the NIST challenge\footnote{\url{https://www.nist.gov/ctl/pscr/open-innovation-prize-challenges/current-and-upcoming-prize-challenges/2020-differential}}, accuracy was measured by the error on the Gini coefficient of incomes and the gender pay gap in (real) demographic data with financial information, when estimated on synthetic data~\cite{Bowen_Snoke_2021}.

The second approach studies whether, for a battery of models, their ranking in terms of accuracy would be the same when trained on synthetic or real data.
This setup assumes that analysts use synthetic data to \textit{select} a model, which is then trained on a real dataset (for better real-world performances).
Crucially, the goal is that model development on synthetic data reflects model development on real data -- when a comparison is made (e.g. between two choices of hyperparameters), it should mirror the comparison on the real data. The utility of the synthetic dataset is then given by how well the performance ranking of models on the synthetic data matches the ranking that would be determined by the real data. This is challenging to measure since, in theory, one would want to ensure that ``all possible methods'' are appropriately ranked, including those which have yet to be developed/discovered. One approach might be to approximate this by comparing a list of representative models~\cite{kddsra}. It might also make sense to expand the list of representative models by incorporating small variations of each model (i.e. by varying the hyperparameters involved). Efficiently computing this for a broad enough class of models would be key, which may require new insights to ensure the class is indeed sufficiently broad.

An important thing to keep in mind when utilising synthetic data in this way is the variability of various models' performances, especially the variability with respect to the real vs.~synthetic data. In particular, if several synthetic datasets (each with decreasing privacy) are going to be used to narrow-down the best methodology, then the process needs to be aware of how the earlier (more private) synthetic datasets will typically create noisier rankings and so the notion of a method being statistically significantly better than another needs to be adjusted accordingly.

\subsubsection{Fidelity-driven evaluation}\label{sec:privfid}
A promise of synthetic data is that it ``looks like'' real data, and can thus be used for a variety of purposes.
From a statistical perspective, the goal that the distribution $\hat{\mathbb pP}$ used to generate synthetic data is \textit{close to} the (unknown) real data distribution $\mathbb{P}$.
Typically, evaluating fidelity in this way involves choosing a distance with which to compare distributions, then evaluating this distance empirically from samples of the real and synthetic datasets.

A simple example is to focus on 1- and 2-way marginals of the data, which can be efficiently computed.
The difference between these marginals can be estimated with a wide range of metrics: total variational distance~\cite{tao2021benchmarking}, correlations and Cramer's V~\cite{tao2021benchmarking}, or classical distances~\cite{Bowen_Snoke_2021}.
These metrics aim to capture whether the synthetic data captures basic properties of the real data, such as histograms of individual attributes and relations between pairs of attributes.

Estimating distributional distances in higher dimensions, capturing relations between several attributes at a time, is challenging.
Researchers have proposed ad hoc metrics, such as comparing the density of the synthetic and empirical distributions over random subsets of $\mathcal{X}$~\cite{Bowen_Snoke_2021}.
Another solution is the \textit{propensity score}~\cite{woo2009global,snoke2018general}, which captures the accuracy of a classifier trained to differentiate real from synthetic data points. Intuitively, if the classifier cannot distinguish the two, then the distribution of synthetic data points must be close to that of real data points (this is the basis of the original GAN framework \cite{gan}).

In the context of generative networks (and specifically, GANs), researchers have proposed measures to evaluate the fidelity of synthetic samples.
Sajjadi et al. proposed metrics of \textit{precision} (the quality of synthetic samples) and \textit{recall} (the diversity of synthetic samples), inspired by common failure modes of GANs~\cite{sajjadi2018assessing}.
The key question that these metrics seek to answer is how do the empirical and synthetic distributions \textit{overlap}: precision (resp. recall) captures how much of the synthetic (resp. real) data falls within the support of real (resp. synthetic) data.
Researchers have proposed extensions of these metrics to increase their robustness to outliers and make them easier to compute~\cite{naeem2020reliable}
and account for the probability distribution rather than just the support~\cite{alaa2021faithful}.

\section{Private Synthetic Data Generation} \label{sec:privsdg}
Thus far, we have focused on privacy in a broader context than synthetic data. This is natural, because most privacy notions apply more generally. However, synthetic data has an additional property that most other (non-synthetic data) outputs do not have -- it resides in the space of the real data. That is, the synthetic data takes values from precisely the same space as the original data. This allows us to use synthetic data in the place of real data, and also to compare the similarity of the synthetic and real data directly. We begin by highlighting some key considerations for private synthetic data generation.

\paragraph{The space of datasets can be very high-dimensional.}
Differential privacy has been conventionally applied to settings in which the dimensionality of the output space is relatively small, such as count queries on rows, classification tasks, etc. Synthetic data generation, on the other hand, gives output in a very high-dimensional space, i.e. the space of datasets (perhaps of a fixed size, $N$). Releasing such a high-dimensional object is challenging under differential privacy because it leads to higher (worst-case) sensitivity of the generating function (i.e.~the algorithm mapping the input dataset to the output dataset).

Because of this increased worst case sensitivity, accurately constructing a dataset under differential privacy (according to some notion of accuracy) is likely to require a large privacy budget. A dataset of 1 million records, each with only 20 features, results in a 20 million dimensional output. Not only is the sensitivity of such a function likely to be high, but even trying to analyse the sensitivity can be an incredibly difficult task. By instead aiming to create a {\em private generator}, one can alleviate some of the difficulties - the complexity of the generator should not need to scale with the number of rows, but instead only with the dimensionality of the data (e.g. the number of columns in tabular data).

\paragraph{Privacy is a property of the Algorithm, {\em not} of the Data.}
An important-to-note property of differential privacy that it is a notion that is concern with probabilistic properties of the generated outputs and not a single realisation/output of the generator. A single output of the generator is is neither private, nor non-private. In a non-synthetic data setting this may be more obvious. If you query the average age of individuals in a dataset, and are given the response `35', without being told how this was computed, you do not know (even with access to the original data) whether the number 35 was computed privately or not. The algorithm that computed this answer could simply be ``always answer 35, regardless of input data'' which clearly reveals no information. The point is there is nothing private or ``unprivate'' about the {\em number} 35.

What {\em is} private (or non-private) is the algorithm that produces the number, or in the case of this report, that produced the synthetic dataset. Crucially, this means that, at least from the perspective of differential privacy, it is {\em meaningless} to talk about the privacy as a property of a concrete synthetic dataset.

\paragraph{Private Data vs. Private Generator.}
The most common approach to generating private synthetic data is to first train a {\em private generator} (e.g.,  \cite{pategan, dpganxie, dpganxinyang, privbayes, dpgm}) on the real data.
This model is then directly sampled from to generate individual datapoints and thus build up an entire synthetic dataset.
Due to the post-processing theorem \cite{dpbook}, if  the procedure used to train the private generator is $\varepsilon$-differentially private ($\varepsilon$-DP), it can be used to generate arbitrarily many synthetic data records without affecting the privacy guarantees.
In fact, the procedure to generate the synthetic dataset (of arbitrary size) is itself $\varepsilon$-DP, which guarantees that individual records in the real data are protected from attacks.

However, training an $\varepsilon-$DP generator to then generate some fixed number of synthetic samples can lead to overly conservative privacy guarantees, and thus lower utility.
Intuitively, DP restricts the amount of information extracted from the real dataset when computing the output (the generator or dataset).
The generator acts as a bottleneck for information: any finite sample from a generator necessarily has less information from the original dataset, and thus typically results in lower utility.

\paragraph{Outliers and Fairness.}
Capturing outliers with private synthetic data is difficult. Outliers are precisely data points with {\em some} uniquely identifying features; this means that ``hiding them in the masses'' becomes impossible. In some scenarios, for example credit card fraud detection, detecting outliers is the goal. In such a setting, private synthetic data is unlikely to provide much utility, as the outliers will necessarily be suppressed by the need for privacy. Indeed, \cite{oprisanu2021measuring} posit that if outliers are to be captured, then an SDG method cannot attain both a high privacy and a high utility.

This has a problematic implication for fairness: minority groups, like outliers, can often end up being under-represented in synthetic data~\cite{pereira2021analysis, ganev2021robin}. Indeed, there is a clear tension between fairness and privacy, with fairness requiring that there is a good utility, even for minority groups, and privacy hiding the contributions of individuals, and thus collectively hiding the contribution of a minority group. Indeed, \cite{cheng2021can} posit that current mechanisms are unable to simultaneously achieve privacy, fairness and utility.

\subsection{Existing Methods and Technologies} \label{sec:privmodels}
Much attention in the machine learning community is being given to the problem of {\em generative modelling}. The goal of generative modelling is to generate samples with similar statistical properties to the available training data. Generative models are a key ingredient in synthetic data generation, but crucially require additional thought beyond their basic capability to ``generate samples'' - the problem of generating samplings from a distribution given training data is under-specified, and can be satisfied by memorising the training data and regurgitating when asked. We defer discussion of generative modelling in general to section \ref{sec:generalgen}, but draw attention to existing {\em privacy-preserving} generative models here.

As already noted in section \ref{sec:dp}, a key ingredient driving many of the deep learning based algorithms is DPSGD \cite{dpsgd} (differentially private stochastic gradient descent), which enables differentially private training of general neural network based architectures. Though not designed specifically for generative models, DPSGD can be applied to both GANs \cite{dpganxie, dpganxinyang, rcgan, dpgm}, and VAEs \cite{dpvae}. Other approaches leverage the popular PATE mechanism \cite{pate1}, which can be applied to black-box models to create a private predictor. This has been used to replace the discriminator in a GAN model with a private PATE model \cite{pategan}, and as a means of passing gradients from discriminator to generator \cite{gpate}. \cite{dpsyn} use a subsample-and-aggregate approach (similar to PATE) alongside differentially private expectation-maximisation (DP-EM) \cite{dpem}.

\sloppy
Other popular approaches involve representing the data in a simple, low-dimensional form, such as using a Bayesian network to represent the data generation process as a series of low-dimensional marginal distributions \cite{privbayes}, or leveraging the classical copula framework \cite{copula} to learn the generation process \cite{dpcopula}. The recent NIST competition\footnote{\url{https://www.nist.gov/ctl/pscr/open-innovation-prize-challenges/current-and-upcoming-prize-challenges/2020-differential}}, was won with an algorithm that learns all 2-way distributions (marginals) in a differentially private way, and then does post-processing to generate data from these (potentially inconsistent) marginals \cite{mckenna2021winning}. See also \cite{zhang2021privsyn} for similar ideas.

\fussy
As should be clear from the proceeding discussion, there are many ways to enforce privacy, even when using the same underlying algorithm (e.g. a GAN). The question of where and how to enforce privacy is very much open. Intuitively, one wants to apply privacy simultaneously: (i) as close as possible to the output (e.g. to the generator of a GAN rather than the discriminator); (ii) wherever the tightest analysis of the privacy can be done (with GANs, the discriminator is typically easier to analyse). With GANs, we see examples of privacy being applied to the discriminator \cite{pategan, dpganxie, dpganxinyang}, and to the generator \cite{gpate}.

\subsection{Partially Synthetic Data} \label{sec:partsdg}
For tabular data, where a user's data is a $n$-tuple of \textit{attributes}  an alternative to the common flavour of SDG is \textit{partially synthetic data}.
In partially synthetic data, the attributes of user records are divided into two categories: \textit{quasi-identifiers} and \textit{sensitive attributes} (formally $\mathcal{X} = \mathcal{X}_\text{quasi} \times \mathcal{X}_\text{sensitive}$).
The quasi-identifiers are assumed to be non-sensitive and can be disclosed unchanged, while the sensitive attributes need to be protected.
Partially synthetic data is obtained by first fitting a statistical model $f:\mathcal{X}_\text{quasi}\rightarrow\mathcal{X}_\text{sensitive}$ on the quasi-identifiers to predict the value of the sensitive attribute, then replacing the sensitive attributes of the data by values produced by the statistical model, a process known as multiple imputation~\cite{rubin1996multiple}.
The synthetic data generated thus contains the quasi-identifiers of all real records.
Researchers have proposed partially SDG models using decision trees, support vector machines, and random forests~\cite{reiter2005using,caiola2010random} (see the \textsf{synthpop} package~\cite{nowok2016synthpop}).

Utility-wise, partially synthetic data have a clear advantage over \textit{fully} synthetic data, as they not require to estimate the full joint distribution $\mathbb{P}(\mathcal{X})$ but only the conditional distribution $\mathbb{P}(\mathcal{X}_\text{sensitive}|\mathcal{X}_\text{quasi})$.
However, some reports suggest that the utility of such data in practice is not particularly appealing, and that they are probably best suited for preliminary data analyses~\cite{loong2013disclosure}, or when combined with restricted access to confidential data~\cite{abowd2001disclosure}.

Privacy-wise, there is the issue that revealing quasi-identifiers might be considered sensitive in some contexts (and this approach is trivially vulnerable to membership inference attacks).
Researchers have shown that current methods are vulnerable to record linkage attacks~\cite{drechsler2008accounting,reiter2009estimating}, where an attacker identifies which record belongs to a target user using the sensitive attributes.
Partially synthetic data generation methods are generally heuristic methods, and do not satisfy guarantees of privacy.
Label DP~\cite{chaudhuri2011sample} is an extension of differential privacy that fits this context well.

\section{De-biased Synthetic Data Generation} \label{sec:debias}
Machine learning models are known to inherit biases from their training data \cite{aisentencing, nlpgenderbias, buolamwini2018gender, amazonrecruit}. De-biasing trained models requires expert knowledge of the model \cite{eqop, kilbertus2017avoiding}, and also an understanding of the different notions of fairness that one may wish to achieve (e.g. fairness through unawareness, demographic parity, conditional fairness \cite{decaf}). An alternative approach that is being explored is to learn to de-bias the dataset itself, thus creating so-called {\em fair data}\cite{feldman2015certifying, kamiran2009classifying, zhang2016causal, calmon2017optimized}.

This data de-biasing can be viewed as a sort of synthetic data generation, in which the synthetic data is the de-biased data. Some approaches take it a step further and explicitly model the problem of data de-biasing as a one of ``ground-up'' generation \cite{decaf, fairgan}. These approaches aim to learn a generative model that itself is fair. In \cite{decaf}, they explore several notions of fairness and, via causal modelling, identify strategies for generating data that satisfy the given notions.

\subsection{Notions of Fairness}
As we have seen, identifying a meaningful, interpretable notion of privacy is difficult, and the same is true for fairness. These are both complex ethical questions that the machine learning community must address sooner rather than later. Unlike privacy, however, obvious notions of fairness {\em do exist}, and are enforceable in a meaningful way\footnote{This contrasts with the fact that, with privacy, the notion of not wanting your data to affect the output {\em at all} is not achievable without simply ignoring your data completely.}. Each notion typically requires some notion of a set of {\em protected attributes}, along which fairness must be ensured (e.g. gender, race).

{\bf Fairness Through Unawareness (FTU)} \cite{ftu1, decaf} requires that the protected attributes, and only the protected attributes, not be used by the predictor. This aligns with the idea that two equally qualified people deserve the same job opportunities, independent of race or gender \cite{decaf}. FTU, however, fails to take into account the effect that protected attributes might have on other unprotected attributes, such as an individual's race resulting in them not being afforded the same educational opportunities as an individual of a different race (and thus resulting in them appearing to be disparately qualified).

\begin{definition}[Fairness Through Unawareness \cite{decaf}]
A predictor $f: \mathcal{X} \to \mathcal{Y}$ is fair if and only if protected attributes $\mathcal{A} \subset \mathcal{X}$ are not explicitly used by $f$ to predict $Y \in \mathcal{Y}$.
\end{definition}

{\bf Demographic Parity (DPa)} \cite{dpar}, instead, requires that a predictors output's not be correlated with the protected attributes. Indeed, with FTU, an attribute that is correlated with the protected attributes can be used as input to a predictor and thus the predictor can indirectly be correlated with the protected attributes (despite not having direct access to them). DPa is a significantly stronger notion of fairness than FTU, which requires adjusting the distributions of {\em all} variables that are correlated with the protected attributes.

\begin{definition}[Demographic Parity \cite{decaf}]
A predictor $f: \mathcal{X} \to \mathcal{Y}$ is fair if and only if protected attributes $\mathcal{A} \subset \mathcal{X}$ are independent of the predictions. That is, given a random variable $X \in \mathcal{X}$, let $A \in \mathcal{A}$ be the components of $X$ that are protected. Then $f$ satisfies demographic parity if and only if $f(X)$ is independent of $A$.
\end{definition}

Under the graphical model approach used in \cite{decaf}, for example, DPa is ensured by deleting all edges that originate from a variable that has a protected attribute anywhere in its causal predecessors. Naturally, such an approach can significantly degrade performance, as many of these variables can be useful predictors of the target. A trade-off is present, in which some fairness may need to be sacrificed for performance, and vice-versa.

\subsection{Limitations of Fair Synthetic Data Generation}
Though one might hope that fair synthetic data would lead directly to fair predictors (which {\em is} the case with private synthetic data and private predictors), this is not the case. Of particular importance to note is the fact that a predictor's fairness is {\em with respect to} a distribution of the features. Indeed, a predictor that is trained on synthetic data may not longer be fair when moved to real data due to a shift in the distribution of the real features. This is partly the reason that \cite{decaf} take such an extreme approach in removing all contaminated features (rather than trying to only remove the influence of the protected attributes on contaminated unprotected ones).

Moreover, a synthetic dataset's fairness is defined through a given predictor. Giving a more general definition for fair synthetic data, and determining whether or not predictors trained on such a synthetic dataset would be fair, are open problems. The hope that an organisation might make a single ``fair'' synthetic dataset for use across an organisation would require advancements in this space, requiring a shift in thinking from fair predictors to fair data.

\subsection{Existing Methods}
As noted above, fair synthetic data generation is a young field. While there is significant amounts of work on creating fair predictors (see e.g. \cite{mehrabi2021survey} for a recent survey), the work on synthetic data for fairness is more limited.

\cite{decaf} take a causal, GAN-based approach, using several GAN networks, along with an assumed known causal graph to learn the generative process of the data. Armed with the causal graph, they then generate synthetic data by selectively dropping edges from the model depending on the notion of fairness being targeted.

\cite{fairgan} use a more indirect approach to ensuring fairness. Again based on a GAN model, they instead opt to introduce an additional loss term that penalises disparity between protected attributes taking different values. This means that the learned model might still generate unfair data, but only if doing so results in an increase in the fidelity of the generated synthetic data. This trade-off is controlled by a hyperparameter than can be increased to more stringently ensure fairness.

\subsection{Evaluating Utility and Fidelity}
Much of the discussion found in Section \ref{sec:utilfidel} can be applied to fair synthetic data. An additional consideration will necessarily be whether or not the bias has been successfully removed, metrics for which can be found in \cite{decaf}. These typically involve a comparison between outputs of a predictor trained on the synthetic data when the input protected attributes are varied.

\section{Data Augmentation} \label{sec:dataaug}
Perhaps the most successful use case (so far) for synthetic data has been for data augmentation - using synthetic data to enlarge datasets with additional samples to use for training \cite{wong2016understanding, timeressemi, timesemi}. This is often referred to as semi-supervised learning. The intuition is that synthetic data can act as a regulariser, reducing variance in the learned downstream model \cite{timeressemi}. Fortunately, there are several good surveys for data augmentation, and so we defer the reader to those for a more thorough background: time-series data \cite{timeseriesreview}; image data \cite{imagereview}.

Of note is the fact that synthetic data driven techniques are more important in domains with less structure (such as with generic tabluar data). In the image domain, there is significant structure that can be exploited to create additional data, such as small rotations, image-flipping, cropping, etc. This structure often does not have a parallel in generic tabular data. As such, augmentation methods driven by synthetic data generators are a promising approach to fill this gap.

\subsection{The Basic Principle}
The key idea driving the use of synthetic data for data augmentation is that of {\em generalisability}. The goal of a model is not just to perform well on its training data, but also on data that has not been seen before by the model \cite{deeplearnrev}. The hope with synthetic data is that one can take the available training data and learn a generative model, such as a GAN \cite{gansemisup}. This generative model will then be able to produce ``realistic'' samples of training data, and, {\em hopefully} these samples will be sufficiently different from the original training data so as to be useful additional data points for the training of the model.

It is particularly important to stress that these samples need to be {\em sufficiently different} from the original data. If these samples are too similar, then they will provide no benefit over using the original data on its own for training. As is hopefully clear by now, this need for dissimilarity is a common theme with synthetic data. In the context of data augmentation, this need is less strict; with privacy and fairness, the dissimilarity is a question of ethics and a failure to satisfy it can have both moral and legal repercussions. With data augmentation, this is unlikely to be the case.

\subsection{What methods are used?}
Depending on the domain in question, a variety of models exist that can perform the task of data augmentation. For the most part, the only real requirement is a generative model that is tailored to the domain in question. In Section \ref{sec:generalgen}, we discuss different types of generative models in more detail. Due to the need for (some level of) dissimilarity, training a generative model and hoping for the best may not be sufficient, however. Instead, one may wish to impose restrictions that enable this dissimilarity. The authors in \cite{badgansemi} show that learning a perfect generator in a GAN leads to poor semi-supervised learning, but that a bad generator performs well, for example. This idea of dissimilarity is often referred to as a need for {\em diversity} in the generated samples. Metrics for measuring this exist, such as the score presented in \cite{alaa2021faithful}.

\section{Generative Modelling - An Overview} \label{sec:generalgen}

\begin{figure}[htp]
    \centering
    \includegraphics[width=0.7\textwidth]{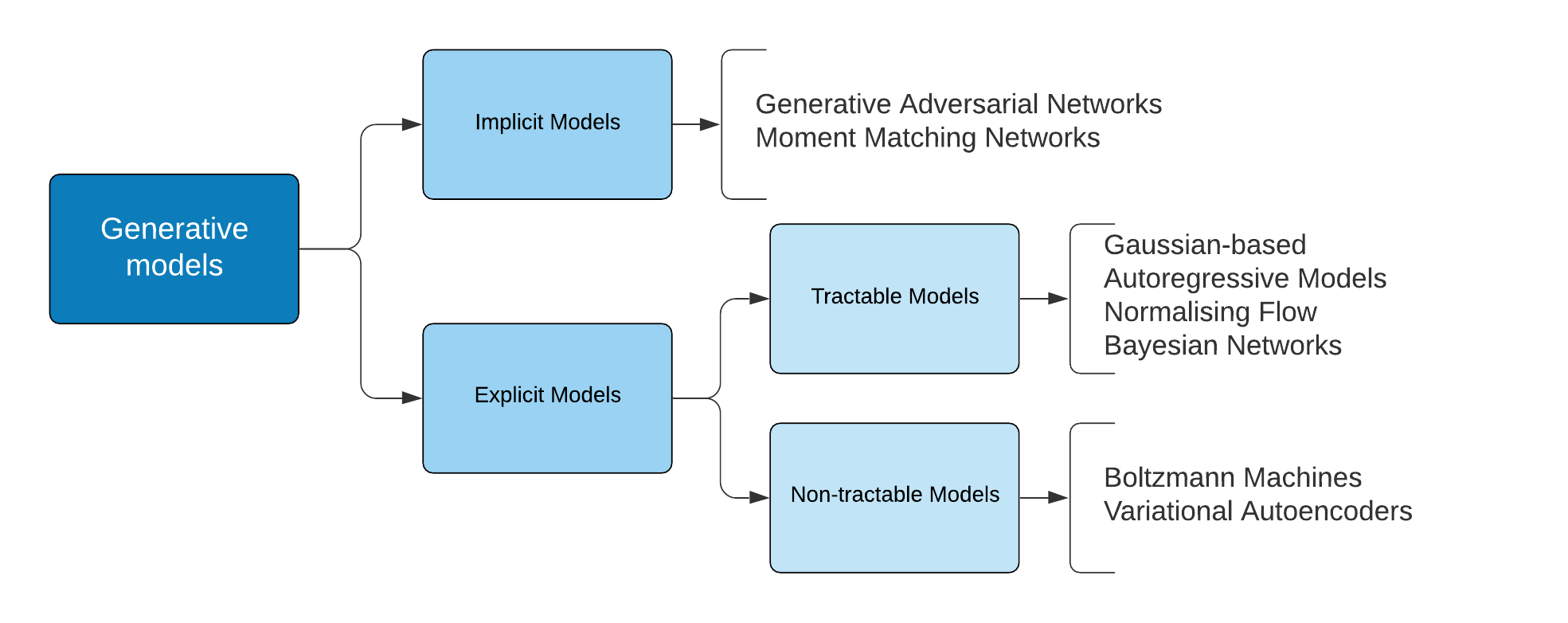}
    \caption{Generative modelling taxonomy based on the maximum likelihood from  \cite{goodfellow_nips_2017} }
    \label{fig:taxonomyGM}
\end{figure}

Though there is a wealth of {\em directed} work on synthetic data generation with a specific goal in mind (such as one of the ones discussed in sections \ref{sec:privacy}, \ref{sec:debias}, and \ref{sec:dataaug}), there is also a lot of work focused on generative modelling [for the sake of generative modelling]. The original GAN paper was introduced as a means to generate fake images, not because they wanted to create private, debiased, or even augmented image datasets, but because the task of generating realistic images was hard and the success was easy to evaluate even for non-experts in the field.  To that end, there are several methodologies that exist that have yet to be applied to a specific problem.

The remainder of this section is more technical than the rest of this manuscript. Those that wish may skip to Section \ref{sec:indmess} without any loss of flow.

These methodologies could be taxonomised as in Figure \ref{fig:taxonomyGM}, as proposed by Goodfellow \cite{goodfellow_nips_2017}, focusing on generative models whose parameters are trained to maximise the likelihood of the original data.
Such methods can be grouped into two main families, where the underlying density function is either explicitly or implicitly defined.
Within explicit models reside statistical methods where new samples are extracted from the distribution arising from the model's definition, which, in turn, must strike a balance between the ability of the model to capture data complexity and to maintain computational tractability \citep{frey_does_1996, breiman2017classification, rasmussen_nite_1999, naivebayes}. 

Non-tractable density functions can be explicitly tackled through deterministic and stochastic approximations. Variational Auto-Encoders (VAEs) \citep{vae} define the loss functions as tractable lower bounds of the non-tractable log-likelihood densities. However, if these deterministic approximations are not carefully calibrated, the model may not converge to the target distribution and consequently generate inconsistent data. On the other hand, stochastic approximations are the basis of Markov chain approaches, where samples are repeatedly drawn after the application of a chosen transition operator. Deep Boltzmann Machines \citep{dbm} are the main representatives of this class, having all neural units composed of random variables, which simultaneously act as inputs and outputs of the closest layers. Such versatility results in difficulties with training, and thus it is preferable to consider the networks as composed of Restricted Boltzmann Machines (RBMs) \citep{rbm}, consisting of only one visible and one latent layer. In a two-pass learning process, RBMs are progressively trained and then globally fine-tuned. Gibbs-sampling is then used to extract synthetic values.

A completely different direction is taken by implicit generative models, which can be thought of as “black-boxes”, where distributions are not explicitly defined but indirectly revealed through sampling. A first example is Generative Stochastic Networks (GSNs), based on Markov chains. In these networks, the distribution is estimated indirectly, employing a parametric transition operator instead of a parametric model. Nonetheless, they are subject to scalability issues, being not efficiently applicable to high-dimensional scenarios. 

Generative Adversarial Networks (GANs) \citep{gan} were designed to be jointly parallel and multi-modal (i.e. capable of simultaneously generating multiple valid outputs for the same input). GANs consist of two networks: the generator and the discriminator. These two networks are trained {\em adversarially}. The generator creates artificial outputs that are passed to the discriminator along with real data. The discriminator is then tasked with identifying which outputs were real, and which were `fake'. The final goal here is to reach equilibrium, in which the generated samples follow the same distribution as the real data. When this happens, the discriminator can do no better than random guessing. Theoretically, GAN generators can perfectly imitate the original distribution provided that the network is sufficiently complex enough and the discriminator is optimal \citep{gan}. In practive, however, training a standard GAN discriminator to optimality can cause convergence issues and zero-gradients for the generator. Attempts to increase model stability include feature matching, label smoothing, and mini-batch discrimination \citep{salimans_improved_2016}. Alternatives to the Jensen-Shannon divergence (JSD) (that is implicitly used by standard GANs) have been investigated, such as the Wasserstein distance in WGAN  \cite{arjovsky_wasserstein_2017}, and the Maximum Mean Discrepancy in MMD-GAN \cite{mmdgan, demmd}. A generalisation of the JSD to f-divergences has also been explored \citep{nowozin_f-gan_2016}.  Unfortunately, no method has proven to be a clear winner \citep{lucic_are_2018}. 
A further risk of GANs lies in mode-collapse, when the generator memorises only a subset of the training information, hence failing to capture the high-level characteristics of the distribution. To tackle this issue, new architectures have been proposed, such as Conditional GANs (CGANs) \citep{cgan}, where additional classification information is provided to both generator and discriminator networks as a form of semi-supervised learning, Deep Convolutional GANs (DCGANs) \cite{dcgan}, where convolutional layers substitute pooling layers and Information maximising GANs (InfoGANs) \citep{chen_infogan_2016} that takes an information-theoretic approach to controlling the generation process. Generative Moment Matching Networks (MMNs) \citep{mmn} constitute another emerging cluster of implicit models, replacing GANs' discriminators with two-sample tests based on kernel maximum mean discrepancy to measure the distance between modelled and target distributions. Although MMNs offer theoretical guarantees, they are currently outperformed by GANs.

Research on generative models is exploding, both in the evolution of existing models as well as in the introduction of new ones. GAN models are the most popular approach, but their implicit nature necessitates the development of trust through the definition and application of rigorous methodologies and metrics, so far demonstrated to be a difficult task.

\begin{sidewaysfigure}[htp]
    \centering{\includegraphics[width=\textwidth]{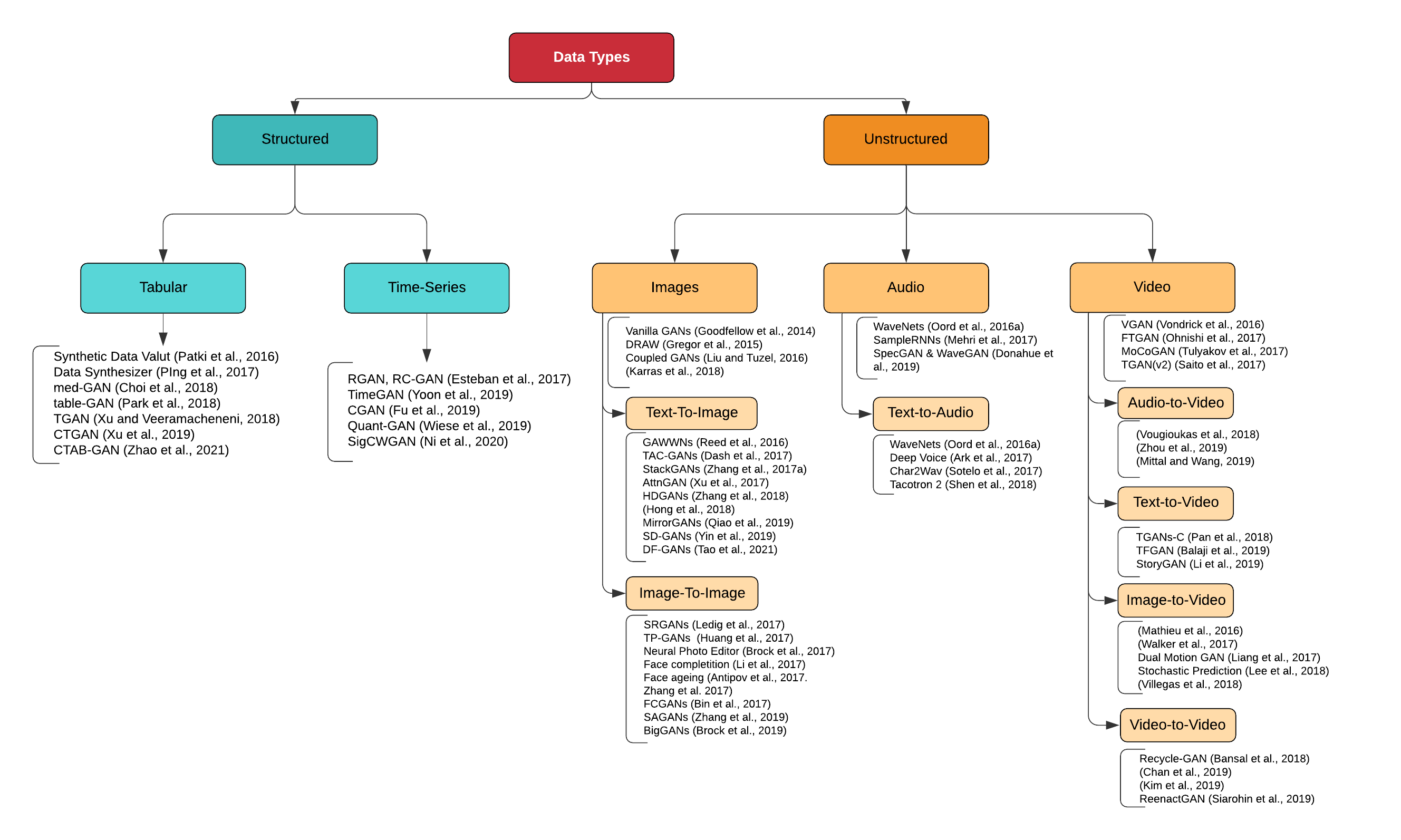}}
    \caption{Generative models for specific data types.}
    \label{fig:data_types}
\end{sidewaysfigure}

\subsection{Existing Methodologies}
The discussion above focused on generative models in the most general sense, without any consideration for the {\em type} of data the model needs to generate. Much work in generative modelling has focused on the image domain with GANs, for example, being originally proposed as an image generation framework. Since their inception, however, many works have looked to generalise GANs to other domains. Below we overview some of the leading generative models that exist for a variety of different data types. Figure \ref{fig:data_types} gives an overview of the taxonomy.

\subsubsection{Tabular data}

Tabular data consists of values stored in rows and columns, whose synthesis requires the simultaneous modelling of distinct column distributions, as well as row-wise and table-wise constraints.
Receiving less attention than image data in the generative domain, tabular data generation still has many obstacles to overcome. Initial generators relied on classifiers, as in the case of inverted decision-trees, vector machines and random forests, which struggle to strike a balance between classifier’s accuracy and the risk of leaking information \citep{caiola_random_2010, drechsler_using_2010, eno_generating_2008}. On the other hand, the application of GANs required the conversion of categorical, discrete columns to a continuous form using auto-encoders or through the decomposition in a variable number of Gaussian modes \citep{choi_generating_2018, xu_modeling_2019, zhao_ctab-gan_2021}. Nonetheless, the independent generation of column values might result in invalid rows, whose semantic correctness requires either the training of additional classifiers \citep{park_data_2018, xu_synthesizing_2018} or techniques based on Gaussian Copulae \citep{patki_synthetic_2016} and Bayesian networks \citep{abowd_how_2008, ping_datasynthesizer_2017,privbayes,gogoshin_synthetic_2020}. 

\subsubsection{Time-series data}

Time-series are series of data points indexed in time order (e.g. electronic health record data containing information about visits to a GP, or higher frequency financial data, such as stock prices).
Historically, they were generated via auto-regressive models \citep{tsay_analysis_2002}, hardly applicable to practical scenarios where stationarity only holds in specific time regions and in the presence of skewed and heavy-tailed distributions. Conversely, most implicit models focused on conditional distributions of future events given the occurrence of past ones, instead of capturing the full joint-law \citep{rcgan, timegan, fu_time_2019, wiese_quant_2019}. Recent developments allow training effort to be optimised by exploiting a reduced feature space, where the data stream is identified by its signature, resulting in a graded sequence of statistics \citep{ni_conditional_2020}.

\subsubsection{Images}
Applications of image synthesis are extremely diverse, ranging from the reconstruction of a damaged or missing region to the improvement of resolution and colour reproduction. The creation of an image consists in choosing a specific colour for each pixel, as the result of an image-to-image transformation or a text-to-image conversion.
Variational Auto-Encoders were somewhat successful \citep{gregor_draw_2015, mansimov_generating_2016}, however, they featured a pixel-wise loss and simple conditioning, as in the case of class labels and image captions. On the contrary, GANs employ a semantic loss which is more aligned with the human visual system as well as being better suited to highly multi-modal outputs, where several valid images could be created  \citep{renault_deep_2019}. Starting from vanilla GANs \citep{gan}, different architectures were proposed to generate plausible images in various datasets, as in the case of human faces \citep{karras_progressive_2018, huang_beyond_2017, brock_neural_2017, li_generative_2017, antipov_face_2017, zhang_age_2017, bin_high-quality_2017}, high-resolution photographs \citep{ledig_photo-realistic_2017, zhang_image_2019, brock_large_2019} and multi-domain images \citep{taigman_unsupervised_2016, liu_coupled_2016}. In text-to-image scenarios, the generation process starts from a brief text description, which is used as additional training information \citep{reed_generative_2016,reed_learning_2016} with the eventual aid of stacked architectures \citep{zhang_stackgan_2017,sharma_chatpainter_2018,zhang_photographic_2018} and attentive, semantic frameworks \citep{xu_attngan_2017,qiao_mirrorgan_2019,hong_inferring_2018} to preserve sentence-level consistency. 

\subsubsection{Audio}
Similar to the generation of time-series, audio signals have a high temporal resolution, requiring representation and synthesis strategies capable of operating efficiently with a large number of dimensions. A significant attempt was the design of WaveNets \citep{oord_wavenet_2016}, arising from the architecture of PixelRNN \citep{oord_pixel_2016} borrowed from the image domain, further evolved by considering the speed difference between the raw audio and the hidden semantic-signal, which is usually many times slower \citep{mehri_samplernn_2017, sotelo_char2wav_2017}. A different approach consisted in the use of spectrograms, i.e. the simultaneous representations of audio signals in time and frequency, which requires lossy assumptions to cope with their non-invertible nature, inevitably reducing the overall quality \citep{shen_natural_2018, donahue_adversarial_2019}. 

\subsubsection{Video}
The synthesis of a video can be considered as the generation of a sequence of images, where the main challenge stems from their inter-dependency and hidden temporal dimension. Unconditional video generation tried to maintain scene and foreground consistencies by separately focusing on objects' motion and RGB frame generation \cite{vondrick_generating_2016, ohnishi_hierarchical_2017, tulyakov_mocogan_2017, saito_temporal_2017}. Conversely, conditional approaches required smaller training datasets and allowed for finer control of modes of distributions, as in the case of audio conditioning for synchronising speech with a talking character \citep{vougioukas_end--end_2018, zhou_talking_2019, mittal_animating_2019}, text conditioning for video generation \citep{pan_create_2018, balaji_conditional_2019, li_storygan_2019}, image conditioning for the prediction of future frames \citep{mathieu_deep_2016, lee_stochastic_2018, villegas_decomposing_2018,walker_pose_2017,liang_dual_2017} and video-to-video for object animation \citep{bansal_recycle-gan_2018,chan_everybody_2019,kim_deep_2018,siarohin_animating_2019}.

\section{Messages from Industry/Start-ups} \label{sec:indmess}
In preparation of this report, the authors interviewed several industry partners and start-ups in the space of synthetic data. In this section we highlight some of the key themes and messages that we received in response to our questions.

{\bf AI itself is in the early stages of mass adoption.} Though serious AI research has been ongoing for a long time now, widespread adoption of AI systems is in its infancy. Synthetic data is a younger field than the  classical AI/ML problems of prediction, clustering, forecasting, etc. and significant research is required to fully benefit from this technology. That said, there is pressure for the adoption of private synthetic data (more than for other technologies) due to a heightening desire from the public for more privacy control.

{\bf Empirical evaluations are necessary.} Though differential privacy is an attractive theoretical notion of privacy, industries struggle to trust it without empirical supporting evidence of privacy. Understanding practical implications of differential privacy (i.e. its susceptibility to attacks, what the parameters actually correspond to) is crucial to enable widespread adoption of private synthetic data.

{\bf Synthetic data cannot wholly replace real data, or can it?} Opinions were more divided on this subject. The impossibility result of Ullman et al. \cite{ullman2011pcps} is a blow against the notion of a completely general-use synthetic dataset. The relaxed result of \cite{boedihardjo2021covariance}, however, indicates it might still be possible in many cases. Several in industry believe that real data should remain at the core of model development, with the final models ultimately being tweaked or even completely re-trained on the real data. Others believe that it will be possible to completely replace real data with synthetic data in the future.

{\bf Synthetic data is about enabling.} The final takeaway is that synthetic data is about enabling processes that would otherwise not be possible, or that perhaps would drain a lot of resources (such as time). Synthetic data could be used to ``access'' data across legislative borders (e.g. in companies with an international presence), or to speed up model development times by allowing model designers access to {\em something} as early as possible. Ultimately, data is very powerful, and synthetic data may allow many more people to tap into its true potential.

\section{Conclusion}
Synthetic data is a promising technology, with a wide variety of applications. For both privacy and fairness, there is a large cost to getting it wrong. The methods that exist today should be implemented with caution, and significantly more research is needed, from a machine learning perspective, but also from a societal perspective, in order to understand properly the methods that exist.

\section*{Acknowledgements}
We would like to thank Accenture, Hazy, HSBC, MOSTLY AI, and the Office for National Statistics for their participation in our interviews.

\bibliographystyle{unsrtnat}
\bibliography{biblio}

\end{document}